\documentclass[sigconf]{acmart}
\AtBeginDocument{%
  \providecommand\BibTeX{{%
    \normalfont B\kern-0.5em{\scshape i\kern-0.25em b}\kern-0.8em\TeX}}}
\copyrightyear{2025}
\acmYear{2025}
\setcopyright{cc}
\setcctype{by}
\acmConference[WWW Companion '25]{Companion Proceedings of the ACM Web Conference 2025}{April 28-May 2, 2025}{Sydney, NSW, Australia}
\acmBooktitle{Companion Proceedings of the ACM Web Conference 2025 (WWW Companion '25), April 28-May 2, 2025, Sydney, NSW, Australia}
\acmDOI{10.1145/3701716.3715250}
\acmISBN{979-8-4007-1331-6/2025/04}
\makeatletter
\gdef\@copyrightpermission{
  \begin{minipage}{0.3\columnwidth}
   \href{https://creativecommons.org/licenses/by/4.0/}{\includegraphics[width=0.90\textwidth]{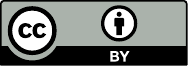}}
  \end{minipage}\hfill
  \begin{minipage}{0.7\columnwidth}
   \href{https://creativecommons.org/licenses/by/4.0/}{This work is licensed under a Creative Commons Attribution International 4.0 License.}
  \end{minipage}
  \vspace{5pt}
}
\makeatother
\usepackage{enumitem}
\usepackage{multirow}
\usepackage{amssymb}
\usepackage{amsfonts}
\usepackage{algorithmic}
\usepackage{graphicx}
\usepackage{makecell}
\usepackage{booktabs} 
\usepackage{amsmath}
\usepackage[normalem]{ulem}
\usepackage{enumitem}
\usepackage{multirow}
\usepackage[nomargin,inline,marginclue,draft]{fixme}
\usepackage{balance}
\usepackage{changepage}
\usepackage{bm}
\usepackage{hyperref}
\usepackage{setspace}
\usepackage{mathrsfs}
\usepackage{ulem}
\usepackage{verbatim}
\usepackage{diagbox}
\usepackage{pdftexcmds}
\usepackage{catchfile}
\usepackage{ifluatex}
\usepackage{ifplatform}
\usepackage{graphicx}
\usepackage{subfigure}
\usepackage{booktabs}
\usepackage[normalem]{ulem}
\useunder{\uline}{\ul}{}
\usepackage{graphicx}
\usepackage{xcolor}
\usepackage{colortbl}
\usepackage[normalem]{ulem}
\useunder{\uline}{\ul}{}
\begin{document}

\title{MRGRP: Empowering Courier Route Prediction in Food Delivery Service with Multi-Relational Graph}

\author{Chang Liu \\ Huan Yan$^{\ast}$ \\ Hongjie Sui}
\affiliation{%
  \institution{Department of Electronic Engineering, BNRist, \\ Tsinghua University, Beijing, China}
  \country{}
}
\email{lc23@mails.tsinghua.edu.cn}
\email{yanhuan@tsinghua.edu.cn}



\author{Haomin Wen}
\affiliation{%
  \institution{Carnegie Mellon University, Pittsburgh, PA, USA}
  \country{}
}

\author{Yuan Yuan}
\affiliation{%
  \institution{Department of Electronic Engineering, BNRist, \\ Tsinghua University, Beijing, China}
  \country{}
}





\author{Yuyang Han \\ Hongsen Liao \\ Xuetao Ding \\ Jinghua Hao}
\affiliation{
  \institution{Meituan, Beijing, China}
  \country{}
  }




\author{Yong Li$^{\ast}$}
\affiliation{%
  \institution{Department of Electronic Engineering, BNRist, \\Tsinghua University, Beijing, China}
  \country{}
}
\email{liyong07@tsinghua.edu.cn}

\thanks{$*$ Huan Yan and Yong Li are corresponding authors.}

\renewcommand{\shortauthors}{Chang Liu et al.}

\begin{abstract}

Instant food delivery has become one of the most popular web services worldwide due to its convenience in our daily lives. A fundamental problem in this service scenario is the accurate prediction of courier routes, which is essential for optimizing task dispatch and thereby improving delivery efficiency. It not only increases the satisfaction of both couriers and users but also drives higher profitability for the platform. The deployed heuristic prediction approach of the platform accounts for only limited human-selected task features and neglects couriers' preferences, resulting in sub-optimal performances. Moreover, existing learning-based methods fail to adequately explore the diverse factors influencing couriers' decision-making behaviors and intricate relationships among factors. To meet the urgent need for a powerful route prediction approach that benefits millions of couriers, users, and the platform itself, we propose a \underline{\textbf{M}}ulti-\underline{\textbf{R}}elational \underline{\textbf{G}}raph-based \underline{\textbf{R}}oute \underline{\textbf{P}}rediction (MRGRP) method, which enables fine-grained modeling of the correlations among tasks influencing couriers' decision-making and achieves accurate prediction. We encode spatial and temporal proximity, along with the pickup-delivery relationships of tasks, into a multi-relational graph, then design a GraphFormer architecture to capture these complex correlations. Furthermore, we introduce a route decoder that leverages information about couriers as well as dynamic distance and time contexts for dynamic prediction. It also utilizes existing route solutions as a reference to find better outcomes. Experimental results demonstrate that our proposed model attains state-of-the-art performance in route prediction on offline data from cities of diverse scales. We then deploy our model on the Meituan Turing online platform, where it significantly surpasses the existing deployed heuristic algorithm and achieves a high route prediction accuracy of 0.819, which is required for the quality and satisfaction of couriers and users in the instant food delivery service. We have open-sourced our code at  \url{https://github.com/tsinghua-fib-lab/MGRoute}.
\end{abstract}

\maketitle

\begin{figure*}[t]
\centering
\vspace{-0.2cm}
\includegraphics*[width=0.85\textwidth]{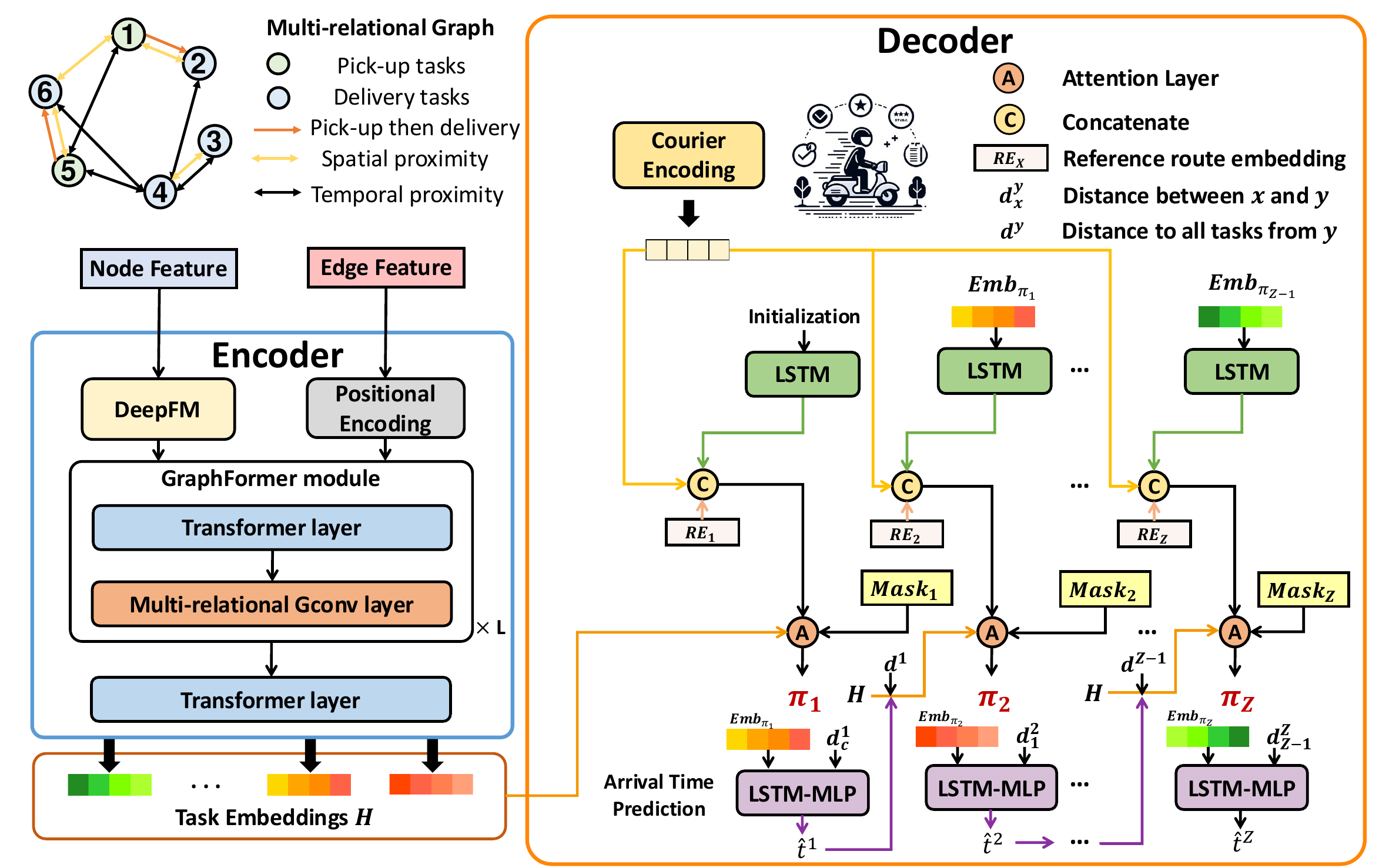}
\vspace{-0.2cm}
\caption{\textbf{The overall framework of MRGRP.}} \label{fig:framework}
\vspace{-0.3cm}
\end{figure*}
\vspace{-0.3cm}
\section{Introduction}\label{Sec:introduction}
Instant food delivery services have become an indispensable aspect of contemporary daily life, primarily due to the convenience they offer in consumption~\cite{lan2020decomposition, DBLP:journals/iotj/AraujoE21, huang2020dynamic}.
After customers submit orders, the service platform communicates the order details to the restaurant and assigns the delivery task to a courier. Once the courier accepts the assignment, they collect the prepared food at the restaurant and subsequently deliver it to the customer's location.
Existing analysis~\cite{gao2021deep,li2018learning,wen2021package} indicate significant deviations between the actual routes of couriers and the recommended routes by the platform. Therefore, one of the fundamental problems in this service scenario is the accurate prediction of courier routes, which is crucial for downstream tasks, including arrival time estimation and task dispatching. 
Effective route prediction facilitates delivery efficiency and task allocation, thereby improving the working experiences of couriers and increasing user retention. 
Ultimately, these advancements contribute to greater profitability for the platform.


Extensive research and numerous initiatives have been undertaken in the field to address this issue. Rule-based methods, such as TimeRank and DisGreedy, predict routes based solely on distance and time, thereby neglecting the intricate features associated with tasks. The heuristic approach TSFH deployed by Meituan, China's largest food delivery service provider, considers only a limited set of human-selected features and overlooks couriers' preferences. Consequently, these methods result in sub-optimal performance and fail to meet the stringent requirements for downstream tasks.
With advancements in learning techniques, significant efforts have been made to leverage these methods to capture diverse task features, thereby enhancing prediction performance. 
For instance, FDNET~\cite{gao2021deep} utilizes a deep architecture to process task features and integrates a combination of LSTM and attention network to predict future routes. Graph2Route~\cite{wen2022graph2route} introduces a dynamic graph neural network to capture task correlations.
Although these methods have demonstrated improved prediction performance, there are still three key challenges that require further consideration:
\begin{itemize}[leftmargin=*,partopsep=0pt,topsep=0pt]
\item \textbf{Inadequate exploration of the multifaceted correlations and dependencies among tasks.} During route selection, it is crucial to simultaneously consider various aspects of task dependencies and correlations, such as spatial-temporal proximity and sequential order. For example, after completing a pick-up task, a courier may face three subsequent delivery tasks: one that is spatially proximate but not urgent, another that is distant but requires immediate delivery, and a third with moderate distance and urgency, corresponding to the same order as the just-picked food. Accurately capturing and managing these diverse and complex factors presents a significant challenge for existing frameworks.
As the number of tasks increases, their interdependencies become increasingly complex, necessitating the development of effective modules to capture these correlations and thereby facilitate route predictions.
\item \textbf{Large solution search space in route prediction.} Route prediction models are usually required to identify the most executable route from a vast feasible solution space that grows exponentially with the number of tasks. Such inherent complexity renders the route prediction problem exceptionally challenging. Consequently, there is a pressing need for the development of advanced frameworks capable of effectively managing the vast solution space of route predictions.
\item \textbf{Inherent dynamism of problem conditions.} The pick-up and delivery process is characterized by an inherent dynamism, where the problem context continuously evolves in response to the courier’s changing location and the passage of time. Such dynamism introduces complexities that cannot be overlooked during route prediction. For instance, the relative distances between tasks shift as the courier moves to a new location. Additionally, certain tasks that may have initially appeared non-urgent can escalate in priority as time progresses, demanding immediate attention. Ignoring this spatial and temporal dynamism can lead to sub-optimal routes and potential delays in service.
\end{itemize}
To deal with the above challenges, we develop a powerful \textbf{\underline{M}}ulti-\textbf{\underline{R}}elational \textbf{\underline{G}}raph-based \textbf{\underline{R}}oute \textbf{\underline{P}}rediction (MRGRP) method with an encoder-decoder architecture. To address the first challenge, we extract and establish multiple relationships among dispatched tasks by considering various factors that influence couriers' decision-making and represent these relationships in the form of a multi-relational graph. Furthermore, we design a multi-relational graph encoder based on the GraphFormer architecture, which proficiently assimilates these intricate correlations among different tasks.
To tackle the second challenge, we design a route decoder that leverages existing competitive route solutions as a reference to guide the prediction process. The decoder utilizes distances and the estimated remaining time of tasks to enhance dynamic route prediction, effectively narrowing the solution space and improving the prediction accuracy.
Additionally, we introduce an auxiliary arrival time prediction procedure that estimates arrival times at each decoding step, thereby enabling the calculation of the remaining time for each task. By further integrating the distance to each task at every decoding step, we construct dynamic features that effectively address the third challenge.                 
In summary, our main contributions are as follows:  
\begin{itemize}[leftmargin=*,partopsep=0pt,topsep=0pt]
\item We propose MRGRP, a novel route prediction approach consisting of an MR-graph encoder designed to capture intricate correlations between tasks within the constructed multi-relational graph and a reference-based route decoder that efficiently narrows the solution space and enables dynamic route decoding.

\item We conduct extensive offline experiments on three real-world datasets to demonstrate the superiority of MRGRP, which improves Location Square Deviation (LSD) by 17.73-20.31\%, and Same Rate (SR) by 3.90-5.96\% over the state-of-the-art baselines. 

\item We deploy our framework on Meituan, China’s largest food delivery platform, and demonstrate that MRGRP outperforms the currently deployed heuristic algorithm, achieving a 31.85\% reduction in LSD and a 7.46\% increase in SR. These improvements are crucial for enhancing the quality of service, satisfaction of both couriers and users and, ultimately, the profitability of the instant food delivery platform of Meituan.
\end{itemize}

\vspace{-0.4cm}
\section{Preliminaries}\label{Sec:perliminaries}
\subsection{Multi-relational Graph}
In food delivery services, there are two types of tasks: pick-up and delivery tasks. 
For a specific courier $c$, his/her unfinished tasks and their relations can be represented with a multi-relational graph $\mathcal{G}^{c}=(\mathcal{V}, \mathcal{E}, \mathbf{X}, \mathbf{E}, \mathcal{R})$, where $\mathcal{V}=\{v_1, v_2, \dots, v_n\}$ denotes the nodes of the graph. 
Each node corresponds to the courier's pick-up or delivery task. $\mathcal{R} = \{r_1, r_2, \dots, r_w\}$ represents $w$ kinds of relations $r$ between tasks. 
$\mathcal{E}=\{(u, v, r) | u, v \in \mathcal{V}, r \in \mathcal{R}\}$ denotes the edge set of the graph, representing the relation type $r$ exist between node $u$ and $v$. $\mathbf{X} \in \mathbb{R}^{n\times d_v}$ denotes node features, where $d_v$ is its dimension. Node features are constructed according to attributes of tasks, such as their geographical location and the predicted finish time.
$\mathbf{E} = \{\mathbf{E}^1, \mathbf{E}^2, \dots, \mathbf{E}^w\}$ represents the edge features for $w$ kinds of relations, where $\mathbf{E}^r \in \mathbb{R}^{n\times n\times d_e^r}$ and $d_e^r$ denotes the edge feature dimension for the relation $r$. Edge features of different relations are collected with the corresponding information between tasks. For example, the edge feature of spatial correlation can include the geographical distance between the two tasks.
\vspace{-0.3cm}
\subsection{Problem Formulation}
Given a courier $c$ and the input graph $\mathcal{G}^{c}=(\mathcal{V}, \mathcal{E}, \mathbf{X}, \mathbf{E}, \mathcal{R})$ including his/her pick-up and delivery tasks and their features as well as correlations, our objective is to derive a function $\mathcal{F}_{\mathcal{P}}$ to predict the route $\boldsymbol{\hat{\pi}}$ of tasks $\mathcal{V}$, which is a permutation of its elements limited by route constraints $\mathcal{P}$:
\begin{equation}
    \mathcal{F}_{\mathcal{P}}(\mathcal{G}^c) = \pi_1, \pi_2, \cdots, \pi_{|\mathcal{V}|}, 
\end{equation}
where $\pi_i \in \{1, 2, \cdots, |\mathcal{V}|\}$ denotes the predicted index of the task finished at step $i$, and $\pi_i \neq \pi_j$ if $i \neq j$. $|\cdot|$ represents the set cardinality. In our problem, we consider a universal pick-up then delivery constraint within $\mathcal{P}$, \textit{i.e.,} the delivery of an order can only be finished after the courier finishes the pick-up task of the same order~\cite{gao2021deep}.

\vspace{-0.3cm}
\section{Methodology}\label{Sec:method}
In this section, we propose a multi-relational graph-based model named MRGRP (short for Multi-Relational Graph-based Route Prediction) to solve the route prediction problem for food delivery services considering the common pick-up then delivery constraint.

Suppose that the courier $c$ has the unfinished task set $\mathcal{V} = \{o_p^1, o_d^1, \dots, o_p^n, o_d^n\}$, where $o$ represents the order from customers, and 
$o_p^i$ and $o_d^i$ denote the pick-up task and the delivery task, respectively. 
We demonstrate the overall architecture of MRGRP in Figure~\ref{fig:framework}, which consists of a multi-relational graph encoder and a reference-based courier route decoder.
Specifically, we first construct a multi-relational graph based on these unfinished tasks, where nodes represent the tasks themselves and edges are defined based on their correlations. The encoder processing on multi-relational graphs transforms task features into highly expressive embeddings. After that, at each decoding step, the decoder recurrently predicts the courier's route selection by utilizing dynamic context information, reference solutions, and the learned task embeddings.
\vspace{-0.3cm}
\subsection{Multi-relational Graph Construction}\label{MRGCons}
Different from the existing method~\cite{wen2022graph2route}, which constructs a task graph containing only a single type of relation that integrates all spatial-temporal correlations between tasks, our approach represents tasks using a multi-relational graph at a more fine-grained level, which can benefit the learning of informative embeddings.

\noindent\textbf{Edge definition.} We define the relations between tasks as follows:
\begin{itemize}[leftmargin=*]
    \item \textbf{Pick-up then delivery.} It is the relation between the pick-up task $o_p^u$ and the delivery task $o_d^u$ of the same order $o^u$. The element $A_{ij}^{pd}$ of its adjacency matrix $A^{pd}$ can be defined as follows:
    \begin{equation}\label{equ:apd}
        A_{ij}^{pd} = \left\{\begin{array}{ll}
            1, & \text{if $i = o_p^u$ and $j = o_d^u$}, \\
            0, & \text{others.}
        \end{array}
        \right.
    \end{equation}
    \item \textbf{Spatial proximity.} It indicates that the geographical distance between node $u, v \in \mathcal{V}$ is relatively close. The adjacency matrix for the spatial proximity relation can be denoted as $A^{sc}$, and its elements are defined as follows:
    \begin{equation}\label{equ:asc}
        A_{ij}^{sc} = A_{ji}^{sc} = \left\{
        \begin{array}{ll}
            1, & \text{if $i$, $j$ are $k$-nearest spatial neighbors,} \\
            0, & \text{others,}
        \end{array}
        \right.
    \end{equation}
    where $k$ is a hyperparameter that can be tuned according to the cardinality of the task set $\mathcal{V}$.
    \item \textbf{Temporal proximity.} We denote the set of pick-up and delivery tasks as $\mathcal{O}_p$ and $\mathcal{O}_d$, respectively.
    The relation indicates the promised pick-up (delivery) time between node $u, v \in \mathcal{O}_p (\mathcal{O}_d)$ is relatively close. The adjacency matrix of the temporal proximity relation can be denoted as $A^{pc}$ ($A^{dc}$), and its element $A_{ij}^{pc}$ ($A_{ij}^{dc}$) is defined as follows:
    \begin{equation}\label{equ:apc}
        A_{ij}^{pc} (A_{ij}^{dc}) = \left\{ \begin{array}{ll}
            1, & \text{if $i$, $j$ are $k$-nearest temporal neighbors,} \\
            0 & \text{others,}
        \end{array}
        \right.
    \end{equation}
\end{itemize}
where $k$ is a hyperparameter.

\noindent\textbf{Node features.} 
Given a task $u \in \mathcal{V}$, its feature comprises multi-domain information, as detailed in Table~\ref{tab:node_feat}.
The features of task $u$ can be classified into categorical and numerical types, which are denoted as $\mathbf{x}_u^c$ and $\mathbf{x}_u^n$, respectively. Please note that our framework is designed to be flexible, allowing the inclusion of additional task features as extra dimensions to further enhance performance.

\vspace{-0.3cm}
\begin{table}[H]
    \centering
    \caption{\textbf{Multi-domain information of tasks.}}
    \vspace{-0.3cm}
    \begin{tabular}{l|p{0.5\linewidth}}
    \hline
       \textbf{Information}  & \textbf{Description} \\
       \hline
         \textbf{Basic info} & Deliver service type, deliver time type, etc.\\
         \hline
         \textbf{Restaurant-related info} & Earliest pick-up time, restaurant GPS coordinate, etc.\\
         \hline
         \textbf{Customer-related info}  & Promised delivery time, customer GPS coordinate, etc.\\
         \hline
         \textbf{Courier-related info}  & Distance to the courier location.\\
         \hline
         \textbf{Area-related info} & Area statistics (the count of tasks per rider, average speed, on-time rates, etc.).\\
         \hline
    \end{tabular}
    \vspace{-0.5cm}
    \label{tab:node_feat}
\end{table}

\noindent \textbf{Edge features.} 
In the multi-relational graph, each edge describes a kind of aforementioned relation between its two ends.
Given an edge $r_{uv} \in \mathcal{E}$, its feature is denoted as $\mathbf{z}_{uv}^r$, where $r$ denotes the relation, and $u$, $v$ represent two edge endpoints, respectively. According to the aforementioned edge definition, $r \in \mathcal{R}=\{1,2,3,4\}$. Their semantics are spatially close, temporally close for pickup tasks, temporally close for delivery tasks, and pick-up then delivery.

      

\vspace{-0.3cm}
\subsection{MR-Graph Encoder}\label{Encoder}
The multi-relational graph (MR-graph) encoder generates highly expressive embeddings for pick-up and delivery tasks, consisting of two key components: i) a node feature encoder and ii) a GraphFormer module~\cite{DBLP:conf/iclr/KipfW17,DBLP:conf/emnlp/MarcheggianiT17,DBLP:conf/esws/SchlichtkrullKB18}.
Each component synergistically enhances the model's capability to capture complex relationships and dependencies inherent in these tasks. Additionally, we integrate an encoding module for the auxiliary task of estimated time of arrival (ETA) prediction, which supports and enhances the primary task of route prediction.

\noindent\textbf{Node feature encoder.} Node features of tasks can be classified into categorical and numerical types. Instead of directly feeding the raw features to downstream modules, we first employ a DeepFM~\cite{guo2017deepfm} framework for feature processing, where sparse categorical features are mapped to dense embeddings and high-order feature interactions are performed adaptively without specially designed feature engineering:
\begin{equation}\label{equ:deepfm}
    \mathbf{x}_u = \text{DeepFM}(\mathbf{x}_u^c, \mathbf{x}_u^n),
\end{equation}
where $\mathbf{x}_u \in d_v$ denotes the output processed feature of task $u$ from DeepFM.
After that, $\mathbf{x}_u$ are then mapped to $d_h$-dimension latent embeddings, formulated as:
\begin{equation}\label{equ:x_u}
    \bar{\mathbf{x}}_u=\sigma(\mathbf{W}_v\mathbf{x}_u + \mathbf{b}_v),
\end{equation}
where $\bar{\mathbf{x}}_u$ constitutes the initial embeddings of task $u$ and can be utilized by downstream modules. $\sigma$ denotes the non-linear activation function, $\mathbf{W}_v\in \mathbb{R}^{d_v \times d_h}$ and $\mathbf{b}_v \in \mathbb{R}^{d_h}$ are trainable parameters.

\begin{figure}[!t]
\centering
\includegraphics[width=0.95\linewidth]{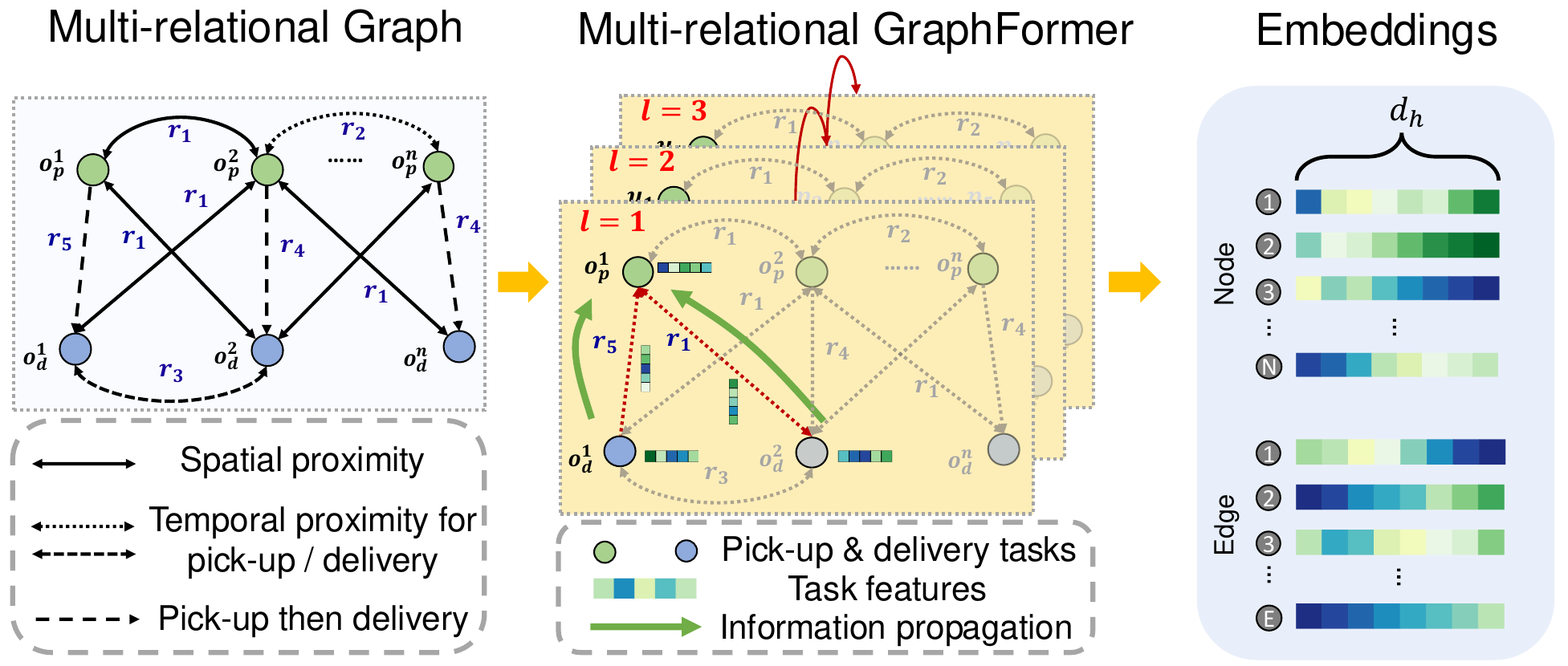}
\vspace{-0.2cm}
\caption{\textbf{The schematic of MRGC layer.}}
\label{fig:mrgraph}
\vspace{-0.5cm}
\end{figure}

\noindent\textbf{GraphFormer module.} 
In the context of the multi-relational graph, traditional graph convolutional networks (GCNs) lack the capabilities to model global dependencies between tasks due to their narrow receptive fields only for connected neighbors and have a risk of over-smoothing in deep convolutional layers. 
Although Transformers are alternatives to perceiving long-range global interactions, they cannot fully leverage the structural knowledge provided by the multi-relational graph. 
To address these limitations, we propose a novel GraphFormer module that effectively leverages the complementary strengths of GCNs and Transformers to encode task features and their correlations. 
It integrates specially designed multi-relational graph convolutional (MRGC) layers between Transformer layers. Specifically, denote the outputs from the node feature encoder as $\mathbf{\bar{X}} \in \mathbb{R}^{|\mathcal{V}|\times d_v}$, we input $\mathbf{\bar{X}}$ to the first Transformer layer, formulated as follows:
\begin{equation} \label{equ:attn}
\mathbf{H}_T^{'(1)} = \text{softmax}(\frac{\mathbf{Q}^{(1)}\mathbf{K}^{(1)T}}{\sqrt{d_{\text{e}}}})\mathbf{V}^{(1)},
\end{equation}
where $\mathbf{Q}^{(1)}=\mathbf{\bar{X}}\mathbf{W}_{q}^{(1)}$, $\mathbf{K}^{(1)}=\mathbf{\bar{X}}\mathbf{W}_{k}^{(1)}$, $\mathbf{V}^{(1)}=\mathbf{\bar{X}}\mathbf{W}_{v}^{(1)}$, and $\mathbf{W}_{q}^{(1)} \in \mathbb{R}^{d_v\times d_e}$, $\mathbf{W}_{k}^{(1)} \in \mathbb{R}^{d_v\times d_e}$, $\mathbf{W}_{v}^{(1)} \in \mathbb{R}^{d_v\times d_e}$ are trainable parameters. To enhance the representation capabilities of the model, we extend it by incorporating multi-head attention, formulated as follows:
\begin{equation}
    \mathbf{H}_T^{(1)}= \text{Concat}(\text{head}_1^{(1)}, \text{head}_2^{(1)}, \cdots, \text{head}_m^{(1)})\mathbf{W}_m^{(1)},
\end{equation}
where we omit the layer superscript for simplicity. Each of head$_i$ is calculated from Equ.~(\ref{equ:attn}), and $\mathbf{W}_m^{(1)} \in \mathbb{R}^{md_e\times d_e}$ is a trainable parameter matrix.
The output of the Transformer layer is subsequently fed to the MRGC layer, which leverages task dependencies provided in the multi-relational graph. The architecture of the MRGC layer is shown in Figure~\ref{fig:mrgraph}.

Specifically, we first map edge features $\mathbf{z}_{uv}^r$ (Section~\ref{MRGCons}) into $d_h$-dimension latent embeddings, denoted as $\bar{\mathbf{z}}_{uv}^r$. A significant distinction between the MRGC layer and the conventional graph convolutional layer lies in the incorporation of relational information. Inspired by existing works~\cite{nickel2015review,vashishth2019composition,liu2022modeling,dettmers2018convolutional}, we employ a composition function $\phi(\cdot)$ to integrate both node and relation embeddings, which can be formulated as:
\begin{equation}
    \mathbf{e}_o = \phi(\mathbf{e}_s, \mathbf{e}_r) = \mathbf{e}_s \odot \mathbf{e}_r,
\end{equation}
where $\phi: \mathbb{R}^{d} \times \mathbb{R}^{d} \rightarrow \mathbb{R}^d$, $\odot$ represents the element-wise product~\cite{DBLP:journals/corr/YangYHGD14a}, $\mathbf{e}_s$, $\mathbf{e}_r$, and $\mathbf{e}_o$ denotes the embedding of the subject, relation, and object, respectively. Therefore, the embeddings of all nodes after the first MRGC layer, $\mathbf{H}_{G}^{(1)}$, can be obtained by aggregating information from its neighboring nodes and relations, formulated as:
\begin{equation}
\label{equ:node_fist}
    \mathbf{h}_{G, v}^{(1)} = \sigma(\sum_{(u,r,v) \in \mathcal{N}_v} \mathbf{W}_{r}^{(1)}\phi(\mathbf{h}_{T, u}^{(1)}, \bar{\mathbf{z}}_{uv}^r)),
\end{equation}
where $r \in \mathcal{R}$ and $\mathcal{N}_v$ is a set of neighbors of node $v$. $\mathbf{h}_{T, u}^{(1)}$ denotes the $u$-th row of $\mathbf{H}_T^{(1)}$.
$\mathbf{W}_r^{(1)}$ corresponds to the trainable weights for relation type $r$. After updating node embeddings, edge embeddings can also be updated as follows:
\begin{equation}
    \mathbf{q}_{uv}^r = \mathbf{W}_{rel}\bar{\mathbf{z}}_{uv}^r,
\end{equation}
where $\mathbf{W}_{rel} \in \mathbb{R}^{d_h \times d_h}$ aims to map all kinds of relations into the same embedding space, and $\mathbf{q}_{uv}^r$ is the updated embedding of relation $r$.
By stacking GraphFormer layers, we are capable of extracting fine-grained information from high-order neighbors and derive node embeddings recursively. The $l$-th GraphFormer layer can be represented as follows:
\begin{align}\label{equ:nodeupdate}
&\mathbf{H}_T^{'(l)} = \text{softmax}(\frac{\mathbf{Q}^{(l)}\mathbf{K}^{(l)T}}{\sqrt{d_{\text{e}}}})\mathbf{V}^{(l)},\\
 &\mathbf{H}_T^{(l)}= \text{Concat}(\text{head}_1^{(l)}, \text{head}_2^{(l)}, \cdots, \text{head}_m^{(l)})\mathbf{W}_m^{(l)},\\
    &\mathbf{h}_{G, v}^{(l)} = \sigma(\sum_{(u,r,v) \in \mathcal{N}_v} \mathbf{W}_{r}^{(l)}\phi(\mathbf{h}_{T, v}^{(l)}, \mathbf{q}_{uv}^{r(l-1)})),
\end{align}
where $l$ denotes the number of layers, $\mathbf{Q}^{(l)}=\mathbf{H}_G^{(l-1)}\mathbf{W}_{q}^{(l)}$, $\mathbf{K}^{(l)}=\mathbf{H}_G^{(l-1)}\mathbf{W}_{k}^{(l)}$, and  $\mathbf{V}^{(l)}=\mathbf{H}_G^{(l-1)}\mathbf{W}_{v}^{(l)}$. $\mathbf{H}_G^{(l)}$ represents the embeddings of all nodes from the $l$-th MRGC layer.
Similarly, edge embeddings after $l$ MRGC layers can be denoted as follows:
\begin{equation}\label{equ:relationupdate}
    \mathbf{q}_{uv}^{r(l)} = \mathbf{W}_{rel}^{(l)}\mathbf{q}_{uv}^{r(l-1)}.
\end{equation}
After propagating $L$ GraphFormer layers, we stack an additional Transformer layer and obtain the final embeddings $\mathbf{H}$ of tasks and $\mathbf{Q}^G$ of relations.

\vspace{-0.3cm}
\subsection{Reference-based Route Decoder} \label{sec:decoder}
Route decoder predicts the next finished task of the courier, with i) a masking mechanism to meet route constraints and avoid unreasonable predictions and ii) a reference-guided dynamic route decoding component for unfinished tasks.

\noindent \textbf{Mask Mechanism.} At each decoding step, we need to mask infeasible actions to meet route constraints and avoid unreasonable results. 
The mask mechanism includes the following masking rules:
\begin{itemize}[leftmargin=*]
\item {\textbf{Mask the tasks outputted at previous decoder steps.}} The output task set before the prediction at decoding step $z$ can be denoted as $\mathcal{V}_z^P$, where the order of tasks has been predicted and we should avoid selecting them repeatedly.
\item{\textbf{Mask the unfinished delivery tasks whose corresponding pick-up task has not been finished.}} It represents the pick-up then delivery route constraint since the food cannot be delivered to the customer if it has not been picked up yet.
\item{\textbf{Mask the tasks that are not $k$-nearest spatial of the last outputted node.}} Couriers typically tend to choose spatially closed tasks to finish next. If a task does not belong to the $k$-nearest spatial neighbors of the most recently outputted node, it is less probable to be the next task the courier wants to complete.
\end{itemize}
During the decoding process, the mask is updated at each step according to the above rules. The mask mechanism allows us to cut off some unreasonable routes and shrink the search space, which contributes to our model's prediction performance.

\noindent \textbf{Reference-based dynamic route decoding.}
For route prediction problems, feasible solution space usually grows exponentially with the number of tasks. Existing works have developed competitive models for route prediction and achieved relatively superior performance~\cite{zhang2019route,gao2021deep,wen2021package,wen2022graph2route,mao2023drl4route,feng2023ilroute,yi2024learning}. Therefore, integrating these solutions during route decoding may provide an effective reference and guide the prediction process. Although the selection of the reference solution is flexible, it should account for the time and storage complexity of the method used to produce the reference solution. In our implementation, we utilize a competitive heuristic method TSFH~\cite{zheng2019two} (Section~\ref{sec:baseline} and Appendix~\ref{app:heuristic}) that operates without requiring training or learnable parameters. It is also deployed on the Meituan Turing Online Platform for route prediction.
Furthermore, the route prediction problem is inherently dynamic, as it evolves with the courier’s shifting location and the progression of time. This necessitates the development of a decoding strategy capable of adapting to constantly changing spatial and temporal conditions. Considering these contexts, we propose our reference-based dynamic task decoder.
The whole decoding process can be formulated as follows:
\begin{align}
    p(\pi|s; \theta)&=\prod_{z=1}^Zp(\pi_z|s, \pi_{1:z-1}; \theta), \\
    &=\prod_{z=1}^Zp(\pi_z|g(f,c ;\theta_e), r_{1:z}, \pi_{1:z-1}; \theta), \label{equ:whole}
\end{align}
where $s$ is the given problem instance. $\theta_e$ denotes the trainable parameters. $f$, $c$ denote the task features and the courier features, respectively. $r$ is the reference route prediction solution. $g(f, c;\theta_e)$ denotes the multi-relational graph encoder. In our parameterization approach, we employ an LSTM module to encode the historically predicted route $\pi_{1:z-1}$. The hidden state at step $z-1$, denoted as $\mathbf{m}_{z-1}$, serves as the encoded representation of the historical route information. Similarly, we employ another LSTM module to embed the former $z$ steps of the reference route $r_{1:z}$, and the hidden state at step $z$, denoted as $\mathbf{r}_{z}$, provides a comprehensive representation of this reference route segment.

Given the significance of distance and the remaining time until the promised arrival in influencing the courier's route decisions, we enrich task embeddings at each decoding step $z$ by incorporating dynamic distance and time features, as formulated below:
\begin{equation} \label{equ:enrich_feat}
    \mathbf{H}^z = \text{Concat}(\mathbf{H}, \mathbf{d}^{z-1}, \mathbf{t}^{z-1}),
\end{equation}
where $\mathbf{H}$ denotes the final embeddings derived from the MR-graph encoder. The term $\mathbf{d}^{z-1}$ represents the distance between each task and the task selected at the previous decoding step $z-1$. Additionally, $\mathbf{t}^{z-1}$ signifies the remaining time until the promised arrival time for all tasks, explicitly defined as $\mathbf{t}^{z-1}=\mathbf{T} - \hat{t}^{z-1}$, where $\mathbf{T}$ denotes the promised arrival time of all tasks, and $\hat{t}^{z-1}$ refers to the estimated time of arrival (ETA) of the task selected at decoding step $z-1$ (detailed in the following part). It is important to note that elements corresponding to masked tasks in $\mathbf{d}^{z-1}$ and $\mathbf{t}^{z-1}$ at each decoding step are set to infinity. 

We subsequently derive the selection probability distribution of candidate tasks, considering the set of masked nodes at decoding step $z$, denoted as $\mathcal{V}_{mask}^z$, based on the aforementioned masking rules:
\begin{equation}\label{equ:uvj2}
        u_v^z = \left\{ 
        \begin{array}{ll}
\sigma(\mathbf{h}_v^z\mathbf{W}_{1})^T\sigma([\textbf{m}_{z-1}, \mathbf{r}_z, r_{ID}^z, \mathbf{e}_{co}]\mathbf{W}_{2})
             & \text{if $v$ $\notin$ $\mathcal{V}_{\text{mask}}^z$} \\
            -\infty, & \text{otherwise,}
        \end{array}
        \right.
\end{equation}
where $\mathbf{h}_v^z$ is the $v$-th row of $\mathbf{H}^z$, \textit{i.e.,} the enriched embedding of task $v$ at decoding step $z$. $r_{ID}^z$ denotes the ID embedding of the route selection at decoding step $z$ from the reference solution. $\mathbf{e}_{co}$ is the embedding of couriers to enable personalized route prediction.
$\mathbf{W}_1, \mathbf{W}_2$ are trainable parameters. $\sigma(\cdot)$ denotes the non-linear activation function GeLU~\cite{hendrycks2016gaussian}. Finally, the output probability of the candidate task can be formulated as follows:
\begin{equation}
     p_v^z = p(\pi_z = v | s, \pi_{1:z-1};\theta) = \text{softmax}(u_v^z).
\end{equation}
Then, the decoder predicts the task to be finished at step $z$:
\begin{equation}
    \pi_z = \text{argmax}_kp_k^z.
\end{equation}

\noindent\textbf{Auxiliary task of arrival time prediction.} We also introduce an auxiliary task of arrival time prediction to augment the route prediction process by providing dynamic time features in Equ.~(\ref{equ:enrich_feat}). Specifically, we employ another LSTM module to model the temporal relations of tasks. At decoding step $z$, the input to the LSTM module is the concatenation of $\mathbf{h}_{\pi_z}$ and $\mathbf{d}_{\pi_{z-1}:\pi_z}$, where $\mathbf{h}_{\pi_z}$ denotes the $\pi_z$-th row of the embedding matrix $\mathbf{H}$ from the MR-graph encoder. $\mathbf{d}_{\pi_{z-1}}^{\pi_z}$ indicates the distance between the previous selected task $\pi_{z-1}$ and the current selected task $\pi_z$. After that, we employ a multilayer perceptron (MLP) to map the hidden embedding of LSTM at step $z$, denoted as $\mathbf{h}_{\text{eta}}^z$, to the prediction result $\hat{t}^{z}$.
\vspace{-0.3cm}
\subsection{Model Training and Prediction}\label{subsec:TAP}
Here we introduce the training and prediction process of MRGRP.

\noindent\textbf{Training.} Since the decoder produces the selection probability distribution of each candidate task, which can be treated as the output of a multi-class classification problem. Hence, we utilize the cross-entropy (CE) loss for the route prediction task, which can be formulated as follows:
\begin{equation}
    \mathcal{L}_{\text{R}} = -\frac{1}{S}\sum_{s\in \mathcal{S}}\sum_{v\in \boldsymbol{\pi}}y_v\text{log}(p(y_v|\theta)),
\end{equation}
where $\mathcal{S}$ is the set of samples, $\mathcal{T}$ represents the set of sampling time steps, and $y_v$ denotes the order of task $v$ in the ground truth of the sample. $p(y_v|\cdot)$ is the predicted probability of task $v$.

For the auxiliary task of arrival time prediction, we introduce the quantile MAE loss~\cite{koenker2005quantile}, formulated as follows:
\begin{equation}
    \mathcal{L}_{\text{ETA}} = \frac{1}{9}\sum_{\gamma=0.1}^{0.9}\sum_{z=1}^{Z}\mathbb{I}_{\hat{t}^z \geq t^z}(1-\gamma)|t^z -\hat{t}^z| + \mathbb{I}_{\hat{t}^z < t^z}\gamma|t^z -\hat{t}^z|,
\end{equation}
where $t^z$ denotes the ground truth of the arrival time. Therefore, the final loss function can be formulated as:
\begin{equation}
\mathcal{L} = \mathcal{L}_{\text{R}} + \alpha\mathcal{L}_{\text{ETA}},
\end{equation}
where $\alpha$ is a hyperparameter controlling the relative weight. In our experiments, we set it as 0.1.
We utilize teacher forcing to facilitate the training of our framework, where we use the actual routes as the input when generating the predicted task at each decoding step. 

\noindent\textbf{Prediction.} MRGRP takes all unfinished tasks as its input and predicts the entire pick-up and delivery routes recurrently. At each decoding step, we use the predicted task generated by the model in the previous decoding step as the input, as shown in Equ.~(\ref{equ:whole}).

\noindent\textbf{Time complexity analysis.} We evaluate the training efficiency of our framework on a single NVIDIA RTX 2080Ti GPU. Each epoch takes approximately 15 minutes, and it takes 20 epochs (5 hours) at maximum for our model to converge among all datasets (see Section~\ref{Sec:experiments} for details). This training duration is acceptable for our industry application, as the trained model is deployed for long-term use without the need for frequent retraining.

\vspace{-0.3cm}
\section{Experiments}\label{Sec:experiments}
Here, we conduct experiments to show the performance of our proposed framework and answer the following research questions:
\begin{itemize}[leftmargin=*,partopsep=0pt,topsep=0pt]
\setlength{\itemsep}{0pt}
\setlength{\parsep}{0pt}
\setlength{\parskip}{0pt}
\item \textbf{RQ1: } How does MRGRP perform compared with different kinds of route prediction models?
\item \textbf{RQ2: } How do different designs of MRGRP affect the model performance?
\item \textbf{RQ3:} How does MRGRP perform in terms of stability when predicting routes of varying lengths?
\end{itemize}
\vspace{-0.2cm}
\subsection{Experimental Settings}
\subsubsection{Datasets.}
We utilize large-scale real-world food delivery service data from Meituan~\footnote{\url{https://www.meituan.com}.}, an influential online food delivery service provider in China, to show the performance of our approach.  Our datasets consist of pick-up and delivery task records of couriers from a month in three different tier cities in China, including Beijing, Chengdu, and Guiyang, which allows us to analyze the robustness and adaptability of our method in both highly developed and less developed urban contexts.
Moreover, such city-level evaluation framework also aligns with existing studies in the field~\cite{wen2022graph2route, feng2023ilroute}.
Each pick-up or delivery task record in the dataset contains various features of the task illustrated in Table~\ref{tab:node_feat} and of the couriers. We aim to predict the future food pick-up and delivery task routes of couriers. The detailed statistics are illustrated in Table~\ref{tab:statistic}. 
\vspace{-0.3cm}
\begin{table}[!h]
\caption{\textbf{Statistics of datasets (ANT: Average number of tasks).}}
\vspace{-0.3cm}
\begin{tabular}{cccc}
\hline
\textbf{Type} & \textbf{City} & \textbf{ANUT} & \textbf{\#Samples} \\ \hline
\multirow{3}{*}{Food P\&D} & Beijing       & 6                      & 859461            \\
                           & Chengdu        & 6                                & 865641            \\
                           & Guiyang       & 5                                & 868881             \\ \hline
\end{tabular}
\label{tab:statistic}
\vspace{-0.5cm}
\end{table}
\begin{table*}[]
\centering
\caption{Overall performance comparison of baselines w.r.t.all metrics. The best results are marked in bold, and we use underlines to denote the second-best results.}
\label{tab:overall_perf}
\vspace{-0.3cm}
\resizebox{\textwidth}{!}{
\begin{tabular}{c|ccccccc|ccccccc|ccccccc}
\hline
\textbf{Dataset} &
  \multicolumn{7}{c|}{\textbf{Beijing}} &
  \multicolumn{7}{c|}{\textbf{Chengdu}} &
  \multicolumn{7}{c}{\textbf{Guiyang}} \\ \hline
\textbf{Model} &
  \textbf{KRC} &
  \textbf{LSD} &
  \textbf{ED} &
  \textbf{SR} &
  \textbf{SR@200} &
  \textbf{HR@1} &
  \textbf{ACC@3} &
  \textbf{KRC} &
  \textbf{LSD} &
  \textbf{ED} &
  \textbf{SR} &
  \textbf{SR@200} &
  \textbf{HR@1} &
  \textbf{ACC@3} &
  \textbf{KRC} &
  \textbf{LSD} &
  \textbf{ED} &
  \textbf{SR} &
  \textbf{SR@200} &
  \textbf{HR@1} &
  \textbf{ACC@3} \\ \hline
TimeRank &
  0.833 &
  1.895 &
  1.211 &
  0.35 &
  0.407 &
  0.604 &
  0.412 &
  0.827 &
  2.135 &
  1.136 &
  0.347 &
  0.399 &
  0.591 &
  0.408 &
  0.835 &
  1.457 &
  1.108 &
  0.375 &
  0.428 &
  0.624 &
  0.434 \\ \hline
DisGreedy &
  0.926 &
  0.777 &
  0.615 &
  0.641 &
  0.756 &
  0.797 &
  0.693 &
  0.929 &
  0.742 &
  0.545 &
  0.66 &
  0.755 &
  0.813 &
  0.713 &
  0.925 &
  0.546 &
  0.552 &
  0.664 &
  0.752 &
  0.813 &
  0.701 \\ \hline
Osquare &
  0.918 &
  0.912 &
  0.718 &
  0.582 &
  0.675 &
  0.748 &
  0.647 &
  0.915 &
  0.914 &
  0.67 &
  0.591 &
  0.672 &
  0.751 &
  0.659 &
  0.915 &
  0.597 &
  0.641 &
  0.61 &
  0.679 &
  0.771 &
  0.663 \\ \hline
TSFH &
  0.949 &
  0.404 &
  0.505 &
  0.705 &
  0.815 &
  0.841 &
  0.757 &
  0.954 &
  0.374 &
  0.431 &
  0.732 &
  0.820 &
  0.860 &
  0.779 &
  0.946 &
  0.327 &
  0.477 &
  0.707 &
  0.788 &
  0.846 &
  0.753 \\ \hline
FDNet &
  0.922 &
  0.901 &
  0.712 &
  0.589 &
  0.681 &
  0.756 &
  0.659 &
  0.924 &
  0.902 &
  0.658 &
  0.596 &
  0.677 &
  0.759 &
  0.664 &
  0.926 &
  0.588 &
  0.629 &
  0.619 &
  0.692 &
  0.78 &
  0.679 \\ \hline
DeepRoute &
  0.928 &
  0.852 &
  0.669 &
  0.602 &
  0.707 &
  0.768 &
  0.671 &
  0.917 &
  1.114 &
  0.673 &
  0.59 &
  0.666 &
  0.76 &
  0.661 &
  0.931 &
  0.526 &
  0.577 &
  0.636 &
  0.701 &
  0.789 &
  0.685 \\ \hline
  JGRM &
  0.937 &
  0.610 &
  0.550 &
  0.676 &
  0.782 &
  0.839 &
  0.736 &
  0.938 &
  0.630 &
  0.510 &
  0.684 &
  0.771 &
  0.841 &
  0.741 &
  0.944 &
  0.406 &
  0.490 &
  0.697 &
  0.776 &
  0.850 &
  0.746 \\ \hline
Graph2Route &
  {\ul 0.953} &
  {\ul 0.344} &
  {\ul 0.478} &
  {\ul 0.721} &
  {\ul 0.824} &
  {\ul 0.852} &
  {\ul 0.769} &
  {\ul 0.958} &
  {\ul 0.325} &
  {\ul 0.407} &
  {\ul 0.751} &
  {\ul 0.831} &
  {\ul 0.877} &
  {\ul 0.801} &
  {\ul 0.952} &
  {\ul 0.282} &
  {\ul 0.444} &
  {\ul 0.743} &
  {\ul 0.815} &
  {\ul 0.87} &
  {\ul 0.787} \\ \hline
DRL4Route &
  0.936 &
  0.608 &
  0.544 &
  0.674 &
  0.785 &
  0.84 &
  0.734 &
  0.935 &
  0.636 &
  0.515 &
  0.682 &
  0.769 &
  0.838 &
  0.739 &
  0.947 &
  0.402 &
  0.486 &
  0.698 &
  0.778 &
  0.852 &
  0.749 \\ \hline
ILRoute &
  0.939 &
  0.612 &
  0.554 &
  0.679 &
  0.78 &
  0.837 &
  0.738 &
  0.94 &
  0.623 &
  0.508 &
  0.686 &
  0.774 &
  0.844 &
  0.743 &
  0.94 &
  0.41 &
  0.498 &
  0.696 &
  0.774 &
  0.848 &
  0.745 \\ \hline
  
\textbf{Ours} &
  \textbf{0.962} &
  \textbf{0.276} &
  \textbf{0.394} &
  \textbf{0.764} &
  \textbf{0.856} &
  \textbf{0.888} &
  \textbf{0.814} &
  \textbf{0.965} &
  \textbf{0.259} &
  \textbf{0.338} &
  \textbf{0.785} &
  \textbf{0.86} &
  \textbf{0.9} &
  \textbf{0.832} &
  \textbf{0.959} &
  \textbf{0.232} &
  \textbf{0.368} &
  \textbf{0.772} &
  \textbf{0.839} &
  \textbf{0.892} &
  \textbf{0.814} \\ \hline
\textbf{Improvement} &
  \textbf{0.94\%} &
  \textbf{19.77\%} &
  \textbf{17.57\%} &
  \textbf{5.96\%} &
  \textbf{3.88\%} &
  \textbf{4.23\%} &
  \textbf{5.85\%} &
  \textbf{0.73\%} &
  \textbf{20.31\%} &
  \textbf{16.95\%} &
  \textbf{4.53\%} &
  \textbf{3.49\%} &
  \textbf{2.62\%} &
  \textbf{3.87\%} &
  \textbf{0.74\%} &
  \textbf{17.73\%} &
  \textbf{17.12\%} &
  \textbf{3.90\%} &
  \textbf{2.94\%} &
  \textbf{2.53\%} &
  \textbf{3.43\%} \\ \hline
\end{tabular}
}
\vspace{-0.3cm}
\end{table*}
\subsubsection{Baselines.}\label{sec:baseline}
We implement representative baseline methods of route prediction for performance comparison, which include rule-based methods, traditional machine learning methods, and state-of-the-art deep models.
\begin{itemize}[leftmargin=*,partopsep=0pt,topsep=0pt]
    \item \textbf{Rule-based models.} i) \textit{DisGreedy}, which selects the nearest accessible task from the courier's current location. ii) \textit{TimeRank}, which selects the most urgent task. iii) \textit{TSFH}~\cite{zhang2019route}, a heuristic route prediction method (Appendix~\ref{app:heuristic}).
    \item \textbf{Machine learning models.} \textit{Osquare}~\cite{zhang2019route}, which trains a LightGBM~\cite{ke2017lightgbm} model to recurrently predict the next finished task at each step and generate the whole route of the courier.
    \item \textbf{Deep models.} i) \textit{FDNet}~\cite{gao2021deep}, which employs an LSTM~\cite{hochreiter1997long} and a Pointer Network~\cite{vinyals2015pointer} for route prediction in food delivery service.
    ii) \textit{DeepRoute}~\cite{wen2021package}, which utilizes a Transformer architecture to predict the route for all unfinished tasks. iii) \textit{Graph2Route}~\cite{wen2022graph2route}, which constructs a spatial-temporal graph from tasks, encoding their correlations.
iv) \textit{JGRM}~\cite{ma2024more}, which learns route embeddings based on self-supervised methods.
v) \textit{DRL4Route}~\cite{mao2023drl4route}, which leverages actor-critic methods to optimize pick-up and delivery routes in dynamic environments. vi) \textit{ILRoute}~\cite{feng2023ilroute}, which models the decision-making process of couriers with imitation learning. 
\end{itemize}
We perform a grid search of hyperparameters of deep learning methods with the validation set to ensure that they can achieve optimal performance for a fair comparison. We conduct each experiment at least five times and report the average results.
\vspace{-0.3cm}
\subsubsection{Evaluation metrics.}
We denote the predicted routes and the ground truths as $\boldsymbol{\hat{\pi}} = [\hat{\pi}_1, \dots, \hat{\pi}_n]$ and $\boldsymbol{\pi} = [\pi_1, \dots, \pi_m]$, respectively, where there could be $m \leq n$ as well as $\text{set}(\boldsymbol{\pi}) \subseteq \text{set}(\boldsymbol{\hat{\pi}})$.
We define $S_{\boldsymbol{\pi}}(i)$ and $S_{\boldsymbol{\hat{\pi}}}(i)$ as the location of the $i$-th task in the ground truth and the predicted route, respectively. We utilize widely accepted metrics to evaluate performances of methods from both local and global perspectives, which include Kendall rank correlation (KRC)~\cite{kendall1938new}, location square deviation (LSD), edit distance (ED)~\cite{nerbonne1999edit}, same rate@k (SR@k), Hit-Rate@k (HR@k), Accuracy@k (ACC@k). Definition details are demonstrated in Appendix~\ref{app:eval_metrics}.

\vspace{-0.3cm}
\subsection{Overall Performance (RQ1)} \label{sec:overall}

We show the overall performances of models \textit{w.r.t.} metrics in global and local perspectives in Table~\ref{tab:overall_perf}, respectively. From the experimental results, we have the following observations:
\begin{itemize}[leftmargin=*,partopsep=0pt,topsep=0pt]
\item \textbf{Our framework effectively achieves the best route prediction performance via MRGRP.} The evaluation results on datasets of three cities demonstrate that our proposed MRGRP can steadily achieve promising performance improvement by a large margin compared with all other baselines.
For example, on Beijing dataset, MRGRP improves the performance of the state-of-the-art baseline (Graph2Route) by $19.77\%$ at LSD, $17.57\%$ at ED, and by $5.96\%$ at SR, $4.23\%$ at HR@1. It also consistently achieves the best route prediction results on the other two datasets. Considering that our model can be deployed to serve millions of couriers in Meituan, a slight increase in the offline experiment metrics can have a significant performance improvement in online scenarios.
\item \textbf{Explicitly integrating relations between tasks facilitates accurate route prediction.} Besides the best performance of MRGRP, we find that Graph2Route~\cite{wen2022graph2route}, which constructs an ST graph from tasks, performs better than other baselines. Since couriers consider multiple factors during route planning, encoding complex correlations between tasks is beneficial to models.
\item \textbf{Distance is an important feature in route prediction tasks.} We find that DisGreedy, a simple strategy to select the nearest tasks in each step, exhibits competitive performances and exceeds some deep models such as FDNet~\cite{gao2021deep} and DeepRoute~\cite{wen2021package}. It underscores the significance of distance as a critical feature of tasks. Regarding distances as ordinary features without explicit modeling may result in their information being conflated with other features, thereby comprising the route prediction performance.
\end{itemize}
\vspace{-0.3cm}
\subsection{Ablation Study (RQ2)}
\begin{figure}[t]
    \centering
    \vspace{-0.5cm}
    \subfigure[LSD (lower better)]
    {\includegraphics[width=0.48\linewidth]{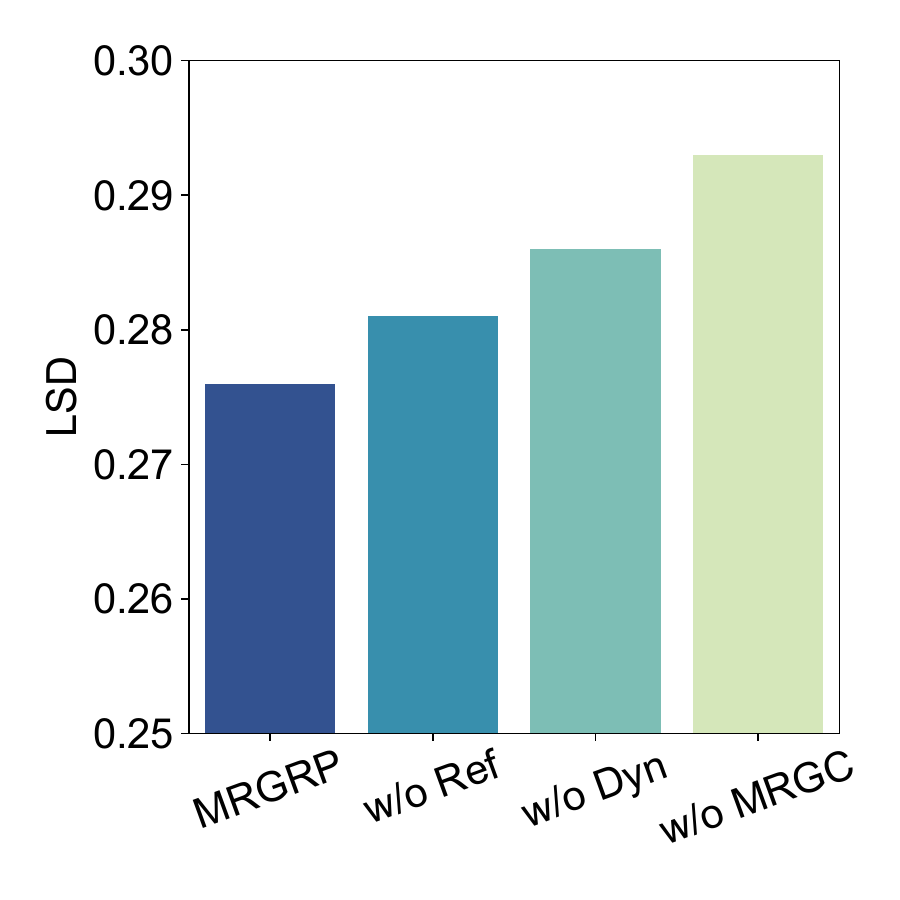} \vspace{-0.2cm}\label{fig:ab:lsd}
    }
    \subfigure[SR (higher better)]{
    \includegraphics[width=0.48\linewidth]{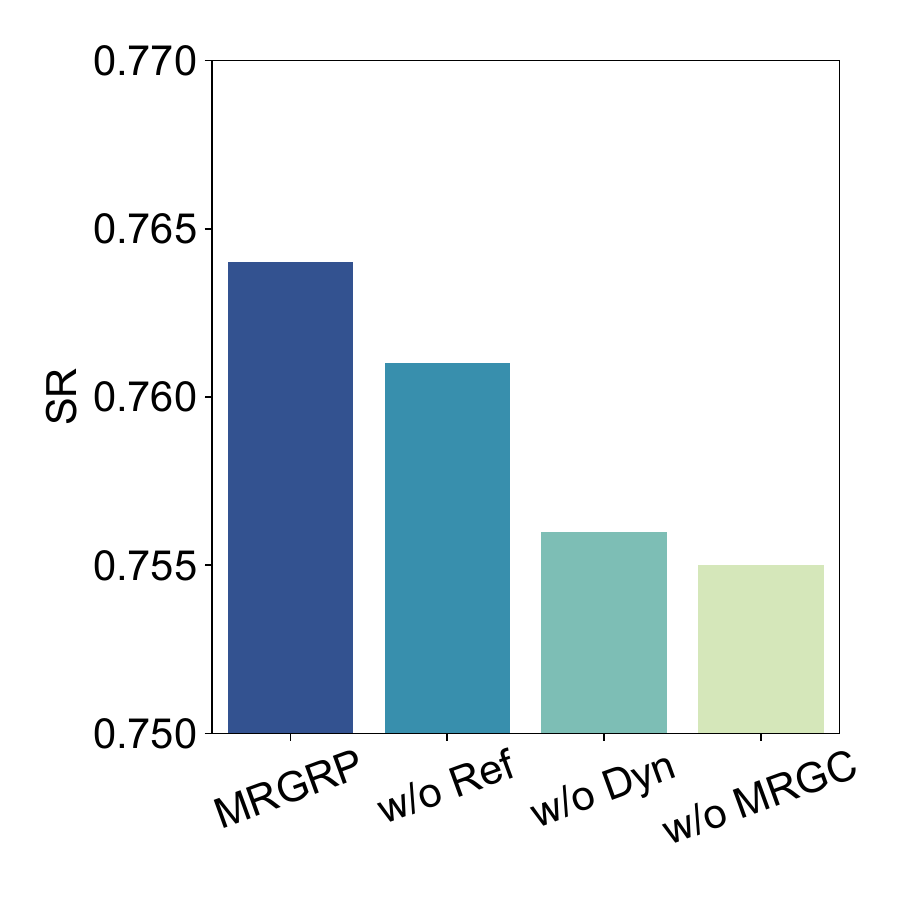}\vspace{-0.2cm}\label{fig:ab:sr}}
    \vspace{-0.6cm}
       \caption{\textbf{Ablation study. w/o Ref/Dyn/MRGC: without reference route solutions/dynamics features during decoding/MRGC layers.}}
    \label{fig:ablation}
    \vspace{-0.7cm}
\end{figure}
We conduct ablation studies to verify the contributions of designed components in MRGRP. 
Specifically, we remove the main components of MRGRP in turn and report the experimental results on Beijing dataset in Figure~\ref{fig:ablation}.
\begin{itemize}[leftmargin=*,partopsep=0pt,topsep=0pt]
\item \textbf{Without MRGC layers.} We exclude MRGC layers from the encoder, retaining only Transformer layers. The results indicate a 0.017 increase in LSD and a 0.010 decrease in SR, underscoring the critical importance of the constructed multi-relational graph.
It demonstrates that leveraging explicit, structured knowledge of task interdependencies significantly benefits the learning of their embeddings and complex correlations.
\item \textbf{Without dynamics features during decoding.} We remove the dynamic distance and time features in Equ.~(\ref{equ:enrich_feat}) at each decoding step and observe a 0.01 increase in LSD and a 0.008 decrease in SR. It underscores the necessity of carefully considering the inherent dynamism of problem conditions in route prediction.
\item \textbf{Without reference route solutions.} We remove the reference task selection at each decoding step. Under this configuration, the LSD of the model increases by 0.005 while its SR decreases by 0.003. These findings highlight the essential role of reference route solutions in guiding the discovery of more optimal routes and effectively narrowing the solution space.
\end{itemize}
\vspace{-0.3cm}
\subsection{Robustness against Route Lengths (RQ3)}
In this section, we investigate the robustness of MRGRP with respect to route lengths, defined by the number of unfinished tasks in each data sample. Samples with 5 or more unfinished tasks are categorized as long-route samples, while those with fewer than 5 unfinished tasks are categorized as short-route samples. We train and evaluate MRGRP, along with other baseline models, on both categories separately to assess their respective performances. Due to the space limit, we demonstrate LSD, ED, SR, and ACC@3 of models on Beijing dataset in Table~\ref{tab:long_short}.
The results indicate that MRGRP consistently outperforms all baseline methods on both long-route and short-route samples. Specifically, it achieves improvements in LSD and ACC@3 by 6.69\% and 7.49\%, respectively, on long-route samples and by 24.53\% and 4.48\%, respectively, on short-route samples. The superior performance can be attributed to the ability of MRGRP to effectively capture and leverage the complex task correlations provided by the multi-relational graph.
\vspace{-0.55cm}
\begin{table}[H]
\centering
\caption{Performances on long and short-route samples.}
\label{tab:long_short}
\vspace{-0.2cm}
\resizebox{\columnwidth}{!}{%
\begin{tabular}{c|cccc|cccc}
\hline
\multirow{2}{*}{\textbf{Model}} & \multicolumn{4}{c|}{\textbf{Long routes}}                 & \multicolumn{4}{c}{\textbf{Short routes}}                 \\ \cline{2-9} 
                                & \textbf{LSD} & \textbf{ED} & \textbf{SR} & \textbf{ACC@3} & \textbf{LSD} & \textbf{ED} & \textbf{SR} & \textbf{ACC@3} \\ \hline
TimeRank                        & 3.484        & 1.744       & 0.204       & 0.273          & 0.572        & 0.793       & 0.513       & 0.52           \\ \hline
DisGreedy                       & 1.387        & 0.885       & 0.541       & 0.632          & 0.301        & 0.404       & 0.753       & 0.74           \\ \hline
Osquare                         & 1.482        & 0.951       & 0.495       & 0.586          & 0.397        & 0.524       & 0.691       & 0.725          \\ \hline
TSFH                            & 0.725        & 0.778       & 0.589       & 0.667          & 0.153        & 0.291       & 0.835       & 0.828          \\ \hline
FDNet                           & 1.785        & 1.18        & 0.413       & 0.505          & 0.235        & 0.371       & 0.777       & 0.772          \\ \hline
DeepRoute                       & 1.717        & 1.023       & 0.437       & 0.521          & 0.384        & 0.433       & 0.698       & 0.695          \\ \hline
JGRM & 1.092 & 0.855 & 0.552 & 0.638 & 0.192 & 0.313 & 0.802 & 0.797 \\ \hline
Graph2Route                     & {\ul 0.523}  & {\ul 0.659} & {\ul 0.642} & {\ul 0.694}    & {\ul 0.159}  & {\ul 0.295} & {\ul 0.835} & {\ul 0.826}    \\ \hline
DRL4Route                       & 1.099        & 0.862       & 0.55        & 0.641          & 0.198        & 0.325       & 0.791       & 0.776          \\ \hline
ILRoute                         & 1.081        & 0.848       & 0.557       & 0.647          & 0.185        & 0.31        & 0.813       & 0.806          \\ \hline
\textbf{MRGRP}       & \textbf{0.488}  & \textbf{0.618}  & \textbf{0.668}  & \textbf{0.746}  & \textbf{0.12}    & \textbf{0.231}   & \textbf{0.864}  & \textbf{0.863}  \\ \hline
\textbf{Improvement} & \textbf{6.69\%} & \textbf{6.22\%} & \textbf{4.05\%} & \textbf{7.49\%} & \textbf{24.53\%} & \textbf{21.69\%} & \textbf{3.47\%} & \textbf{4.48\%} \\ \hline
\end{tabular}%
}
\vspace{-0.5cm}
\end{table}
\vspace{-0.2cm}
\section{Online Deployed Evaluation} 
We have incorporated and deployed MRGRP to the Meituan Turing Platform for more than a month, which operates one of the largest food delivery systems in China and processes millions of daily orders to predict the routes of couriers. We train MRGRP with 10 million real courier routes randomly sampled from country-level data and test its performance on randomly sampled 50k routes. We also sample test data according to their lengths and evaluate the performance of MRGRP. The results, as presented in Table~\ref{tab:online_evaluation}, demonstrate the superiority of our approach compared to the already deployed TSFH~\cite{zheng2019two} on the platform. We find that MRGRP achieves significant improvements on routes with random lengths across various metrics, yielding 31.85\% and 7.46\% improvement in LSD and SR, respectively. MRGRP exhibits an 11.52\% improvement in SR for datasets with long routes and a 4.75\% improvement for short routes. These results underscore the effectiveness of MRGRP in route prediction across diverse real-world scenarios. When utilizing our predicted routes for arrival time estimation, we observe a decrease in its MAE by 14.61\%, 16.07\%, and 12.98\% on random, long, and short routes, respectively. These improvements significantly benefit delivery efficiency, user satisfaction and retention of the instant food delivery service provided by Meituan.
\vspace{-0.3cm}
\vspace{-0.1cm}
\begin{table}[h]
\centering
\caption{Results of online deployed evaluation.}
\vspace{-0.3cm}
\label{tab:online_evaluation}
\resizebox{\columnwidth}{!}{%
\begin{tabular}{c|cc|cc|cc}
\hline
\multirow{2}{*}{\textbf{Model}} & \multicolumn{2}{c|}{\textbf{Random routes}} & \multicolumn{2}{c|}{\textbf{Long routes}} & \multicolumn{2}{c}{\textbf{Short routes}} \\ \cline{2-7} 
               & \textbf{LSD}   & \textbf{SR}    & \textbf{LSD}   & \textbf{SR}    & \textbf{LSD}   & \textbf{SR}    \\ \hline
TSFH      & 0.416          & 0.762          & 0.647          & 0.692          & 0.154          & 0.839          \\ \hline
\textbf{MRGRP} & \textbf{0.284} & \textbf{0.819} & \textbf{0.419} & \textbf{0.772} & \textbf{0.134} & \textbf{0.879} \\ \hline
\textbf{Improvement}            & \textbf{31.85\%}      & \textbf{7.46\%}     & \textbf{35.22\%}    & \textbf{11.52\%}    & \textbf{13.20\%}     & \textbf{4.75\%}    \\ \hline
\end{tabular}%
}
\vspace{-0.5cm}
\end{table}


\section{Related Works}\label{Sec:related_work}
Related works are classified into two groups: next location prediction and route prediction.

\noindent\textbf{Next Location Prediction}.
This task is to predict which locations an individual will visit in future time slots. 
Many researchers proposed deep learning-based approaches to address this task~\cite{petzold2005next,yao2017serm,HST2018Kong,feng2018deepmove,zhao2020go,chen2020cem,li2020cooperative,yang2020location,sun2022tcsa,hong2022you,yan2025generative}.
For example, by incorporating spatial-temporal influence, HST-LSTM~\cite{HST2018Kong} utilizes LSTM to forecast the movement of users in the upcoming minutes or hours.
CTLE~\cite{lin2021pre} proposed a location embedding model that considers both contextual and temporal information in trajectories.
TCSA-Net~\cite{sun2022tcsa} developed a temporal-context-based self-attention network to capture both long-term and short-term preferences of user movements. A common of these works is to infer the possible location from the whole location set in the data. 
Different from them, our work focuses on selecting the location of the pick-up or delivery task from a given set containing the locations of unfinished tasks.
Meanwhile, since the pick-up and delivery tasks in food delivery services have strict spatial and temporal constraints, it is hard to use the above-mentioned methods in our work directly.

\noindent\textbf{Route Prediction}.
Route prediction is previously formulated as a path-finding problem and addressed by minimizing the search space with heuristic algorithms~\cite{wei2012constructing,luo2013finding}. Recently, more researchers have studied the problem of route prediction with learning approaches in this field~\cite{zhang2019route,gao2021deep,wen2021package,wen2022deeproute+,wen2022graph2route, feng2023ilroute, mao2023drl4route}.
For example,~\cite{zhang2019route} designed a machine-learning model to extract the features of decision-making psychology to predict couriers' future routes. 
DeepRoute~\cite{wen2021package} proposed a novel deep model to predict couriers' routes, where a transformer encoder is utilized to capture the spatio-temporal dependencies between unpick-up packages, and an attention-based decoder is adopted to infer the pick-up order of the unfinished tasks recurrently. 
Graph2Route~\cite{wen2022graph2route} designed a spatio-temporal graph neural network to explore the spatio-temporal correlations between different tasks and utilized the graph structure and couriers' profile to decode future routes.
ILRoute~\cite{feng2023ilroute} utilized a graph-based imitation learning framework that models the decision-making process of food delivery couriers.
However, these works cannot fully explore how factors related to pickup and delivery decision-making affect courier behavior. Unlike them, we propose to build a multi-relational graph to consider these factors and employ the effective reference solution and dynamical environment contexts to improve the accuracy of route prediction.

\vspace{-0.4cm}
\section{Conclusion}
We propose a powerful approach, MRGRP, to predict the delivery routes. It utilizes a specifically designed MR-graph encoder to capture intricate correlations of tasks.
The decoder leverages existing route solutions as references and couriers' personalized features during predictions while narrowing the solution space. Additionally, it integrates dynamic distance and time contexts, considering the inherent dynamism of delivery conditions. When deployed on the Meituan Turing Platform, it significantly improves the currently used algorithm, substantially enhancing delivery efficiency, user satisfaction, and retention.
\vspace{-0.3cm}
\section*{Acknowledgment}
This work is supported in part by National Natural Science Foundation of China under 62272260, 62476152, U24B20180.



\bibliographystyle{ACM-Reference-Format}
\bibliography{main}

\appendix
\vspace{-0.2cm}
\section{Appendix}
In this appendix, we introduce additional details of the baseline deployed algorithms, evaluation metrics, and experimental results.
\vspace{-0.2cm}
\subsection{TSFH Algorithm}\label{app:heuristic}
The heuristic algorithm TSFH~\cite{zheng2019two} operates through a two-step process involving initialization and local search strategies. 
This algorithm optimizes route scheduling by considering both the distance between tasks and their estimated time of arrival (ETA).
Each order consists of a pick-up task and a delivery task.
During the initialization phase, the algorithm sorts all orders according to their ETA and plans the first order according to the pick-up then delivery constraint.
For subsequent orders, the algorithm greedily inserts pick-up and delivery tasks into the route, aiming to minimize delays and route length.
To speed up the initialization process, the algorithm utilizes geographic information to cluster pick-up and delivery tasks, considering that groups of restaurants and customers are often geographically proximate. 
This clustering method effectively reduces sub-optimal insertion attempts, ensuring that each task is primarily inserted within its respective cluster.
After the initialization stage, the algorithm applies a local search to further refine the solution by exploring its neighborhood.
During this phase, the algorithm iteratively identifies delivery tasks with significant time margins. Specifically, tasks with an ETA more than five minutes earlier than the promised delivery time are adjusted backward to optimize the schedule. Conversely, tasks with delays exceeding five minutes are moved forward to mitigate lateness. The solution is continuously updated whenever a better route is discovered through the local search.
\vspace{-0.3cm}
\subsection{Details of Evaluation Metrics} \label{app:eval_metrics}
\noindent \textbf{KRC}: Kendall Rank Correlation (KRC)~\cite{kendall1938new} is a statistical measurement to the ordinal association between two sequences. For a task pair $(u, v)$, it is regarded as a concordant pair when both $S_{\boldsymbol{\hat{\pi}}}(u) > S_{\boldsymbol{\hat{\pi}}}(v)$ and $S_{\boldsymbol{{\pi}}}(u) > S_{{\hat{\pi}}}(v)$, or $S_{\boldsymbol{\hat{\pi}}}(u) < S_{\boldsymbol{\hat{\pi}}}(v)$ and $S_{\boldsymbol{{\pi}}}(u) < S_{{\hat{\pi}}}(v)$. Otherwise, it is regarded as a discordant pair. As mentioned before, since the tasks in the predicted route may not exist in the ground truth, we categorize them into two sets, consisting of i) tasks in the ground truth ($\mathcal{V}_{in} = \{\hat{\pi}_i | \hat{\pi}_i \in \boldsymbol{\pi}\}$) and ii) tasks not in the ground truth ($\mathcal{V}_{not}=\{\hat{\pi}_i | \hat{\pi}_i \notin \boldsymbol{\pi}\}$). The order of tasks in $\mathcal{V}_{in}$ are easy to obtain, and we can also know that the order of tasks in $\mathcal{V}_{in}$ are ahead of that in $\mathcal{V}_{out}$. Consistent with the existing work~\cite{wen2022graph2route}, we compare task pairs $\{(u,v)|u, v \in \mathcal{V}_{in}, u \neq v\} \cup \{(u,v)|u\in \mathcal{V}_{in}, v \in \mathcal{V}_{out}\}$. The KRC is defined as follows:
\begin{equation}
    \text{KRC}=\frac{N_c - N_d}{N_c + N_d},
\end{equation}
where $N_c$ and $N_d$ denotes the number of concordant and discordant task pairs, respectively.

\par\noindent\textbf{LSD}: The Location Square Deviation (LSD) 
is used to measure the degree that the predicted route deviates from the ground truth, which can be formulated as:
\begin{equation}
    \text{LSD} = \frac{1}{m}\sum_{i=1}^m(S_{\boldsymbol{\pi}}(\pi_i) - S_{\boldsymbol{\hat{\pi}}}(\pi_i))^2.\\
\end{equation}
\par\noindent\textbf{ED}: Edit Distance~\cite{nerbonne1999edit} is the metric to measure the dissimilarity of sequences via counting the minimum required steps to transform one sequence to another.
\par\noindent\textbf{Same rate}@k (\textbf{SR}@k): 
It quantifies the relaxed consistency rate of the predicted route and the ground truth.
In particular, we measure it using the ratio of the common prefix length of two sequences to the length of the ground truth. $k$ represents a specified distance. If the distance between the predicted node and the node in the ground truth is less than $k$ meters, we consider the model to provide accurate prediction results at this step and count it into the common prefix. The metrics are designed to relax the evaluation of prediction results and are reasonable in real-world scenarios. For example, the courier can pick up two tasks at the same merchant; it is meaningless to evaluate the order of two tasks strictly. In our experiments, we use SR@1 and SR@200 as the evaluation metrics.
\par\noindent\textbf{HR}@k: Hit-Rate@k is used to measure the similarity between top-k items of two sequences, which can be formulated as follows:
\begin{equation}
    \text{HR}@k = \frac{\boldsymbol{\hat{\pi}_{[1:k]}} \cap \boldsymbol{{\pi}_{[1:k]}}}{k}.
\end{equation}
\par\noindent\textbf{ACC}@k: ACC@k is used to calculate the local similarity of two sequences. It measures whether the first k tasks of the predicted route are exactly the same as those of the ground truth, which can be formulated as follows:
\begin{equation}
    \text{ACC}@k = \prod_{i=0}^k \mathbb{I}({\hat{\pi}_i}, \pi_i),
\end{equation}
where $\mathbb{I}({\hat{\pi}_i}, \pi_i) = 1$ when $\hat{\pi}_i = \pi_i$, otherwise $\mathbb{I}({\hat{\pi}_i}, \pi_i) = 0$.
\vspace{-0.3cm}
\subsection{Further Experimental Results}
\subsubsection{Case Study}
\begin{figure}[t]
    \centering
    \includegraphics[width=\linewidth]{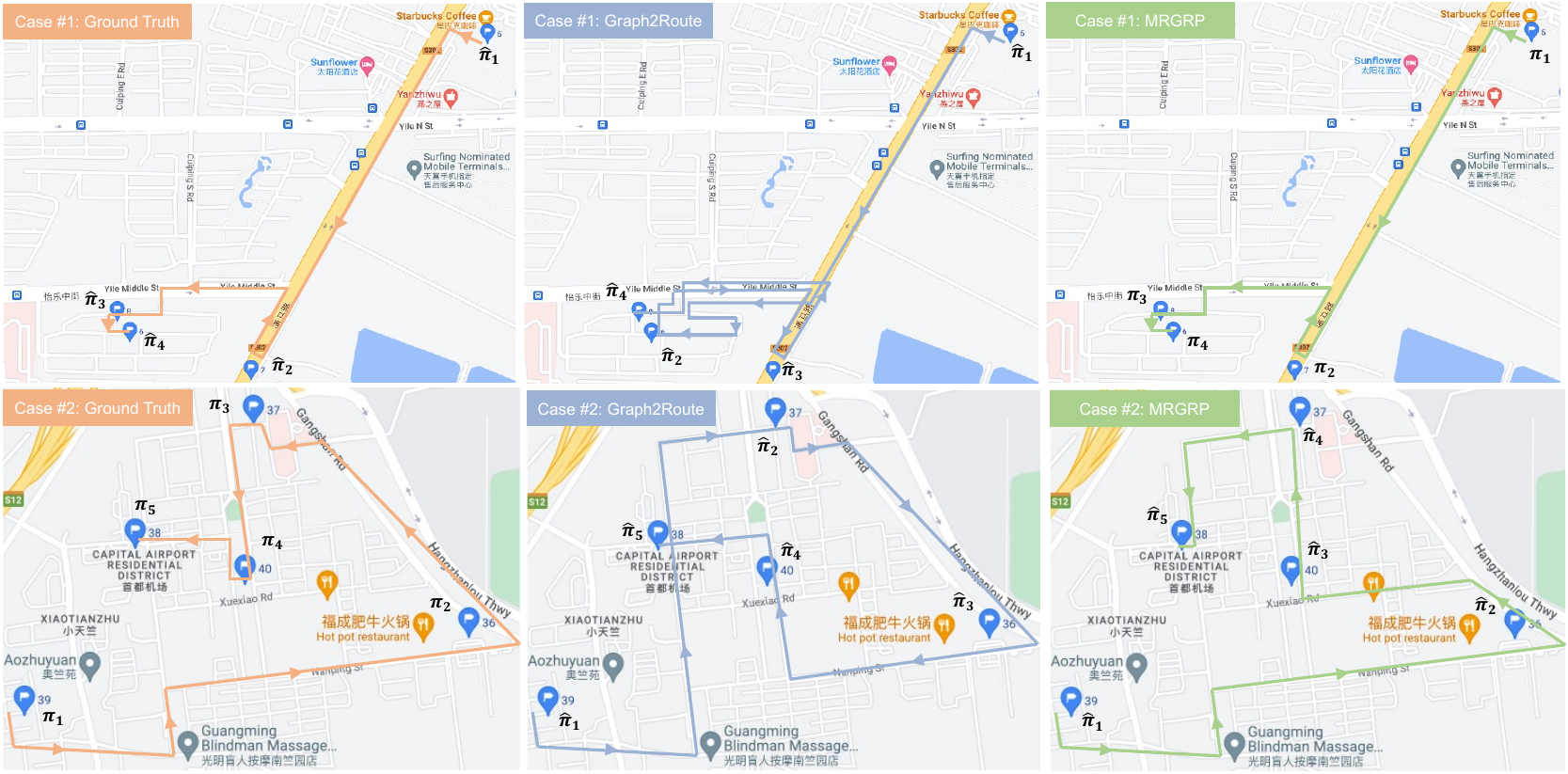}
    \vspace{-0.8cm}
    \caption{\textbf{Case study. Orange, blue, and green lines correspond to real routes, routes predicted by Graph2Route, and MRGRP.}}
    \label{fig:case_merge}
    \vspace{-0.65cm}
\end{figure}

To provide an intuitive analysis of MRGRP, we use empirical route prediction cases to illustrate its performance and advantages over the state-of-the-art model (Graph2Route), which are shown in Figure~\ref{fig:case_merge}. 
For each case, we illustrate the real routes of a courier, routes predicted by Graph2Route and MRGRP. 

\begin{itemize}[leftmargin=*,partopsep=0pt,topsep=0pt]
\item \textbf{Case \#1:} 
We find that MRGRP predicts a more reasonable and accurate route compared to the baseline Graph2Route, which is also consistent with the ground truth. There are four pick-up and delivery tasks for two orders denoted by $5\sim 8$ on the map, where $5$ and $7$ are pick-up tasks, and $6$ and $8$ are corresponding delivery tasks. 
The prediction of MRGRP is identical to the actual route of the courier and is an efficient solution. 
However, the route predicted by Graph2Route is unreasonable; the courier needs to enter and depart the same residential area twice, and the length of the route is much longer than the actual route.

\item \textbf{Case \#2:} 
In this case, there are five pick-up and delivery tasks for three orders denoted by $36\sim 40$ on the map, where $37$ and $39$ are pick-up tasks and $36, 38$, and $39$ are delivery tasks. 
Despite the fact that both MRGRP and Graph2Route fail to accurately predict the entire path, we find that the predicted route of MRGRP is also reasonable,
and the route length does not differ significantly from the ground truth, which is less likely to cause serious harm to downstream tasks such as arrival time prediction and order dispatching. 
However, the route predicted by Graph2Route deviates much from the ground truth. For example, it predicts the task $37$ in the second step, which is the farthest from the previous predicted task $39$. Therefore, the prediction results are not reliable and cannot contribute much to the other tasks.
\end{itemize}
\vspace{-0.2cm}
\subsubsection{Hyperparameter Study}
We performed comprehensive hyperparameter tuning to identify the optimal configuration for model performance. Our study revealed that setting the number of MRGC layers to 3, the task embedding size to 256, and the relative weight of the loss functions to 0.1 yielded the best results. Due to space constraints, detailed results are provided in our GitHub repository~\url{https://github.com/tsinghua-fib-lab/MGRoute}.
\end{document}